\documentclass[letterpaper, 10 pt, conference]{ieeeconf}  

\IEEEoverridecommandlockouts                              

\usepackage{amsmath} 
\usepackage{amssymb}  
\usepackage{graphicx}
\usepackage{algorithm}
\usepackage{algpseudocode}

\title{\LARGE \bf
Planning for Tabletop Object Rearrangement
}

\author{Jiaming Hu \and Jan Szczekulski \and Sudhansh Peddabomma \and Henrik I. Christensen
\thanks{Contextual Robotics Institute, UC San Diego, La Jolla, CA 92093, USA}
}

\begin{document}

\maketitle
\thispagestyle{empty}
\pagestyle{empty}

\begin{abstract}

Finding an high-quality solution for the tabletop object rearrangement planning is a challenging problem. Compared to determining a goal arrangement~\cite{tidy}, rearrangement planning is challenging due to the dependencies between objects and the buffer capacity available to hold objects. Although~\cite{orla} has proposed an A* based searching strategy with lazy evaluation for the high-quality solution, it is not scalable, with the success rate decreasing as the number of objects increases. 
To overcome this limitation, we propose an enhanced A*-based algorithm that improves state representation and employs incremental goal attempts with lazy evaluation at each iteration. This approach aims to enhance scalability while maintaining solution quality. Our evaluation demonstrates that our algorithm can provide superior solutions compared to~\cite{orla}, in a shorter time, for both stationary and mobile robots.

\end{abstract}

\section{INTRODUCTION}

Tabletop Object Rearrangement with Overhand Grasps (TORO) in bounded workspaces is a challenging manipulation task in robotics, with applications ranging from automated assembly lines to household assistance~\cite{rearrangementsur}. While many approaches can generate feasible rearrangement solutions~\cite{trlb}, optimizing these solutions to minimize travel costs remains a challenging problem. This complexity arises particularly from the need to consider dependencies between objects to avoid collisions. Consequently, developing an algorithm that can efficiently search for high-quality plans with minimal travel cost is essential. This work aims to explore such an algorithm.

To find high-quality rearrangement plans with minimal cost, the Object Rearrangement with Lazy A* (ORLA*) algorithm~\cite{orla} was introduced as a state-of-the-art method. ORLA* extends the A* algorithm for tabletop rearrangement planning by treating object arrangements as states and using pick-and-place actions to transition between them. To reduce computation times, ORLA* employs lazy evaluation, which avoids unnecessary buffer allocations and enhances efficiency. ORLA* still faces a challenge in terms of scalability. Like A*, it only terminates the search upon finding a solution or timing out, which implies exponential increases in computation time as the number of objects grows. This scalability issue makes it difficult for ORLA* to provide high-quality solutions in reasonable time for complex scenes.

\begin{figure}
    \centering
    \includegraphics[width=0.5\textwidth]{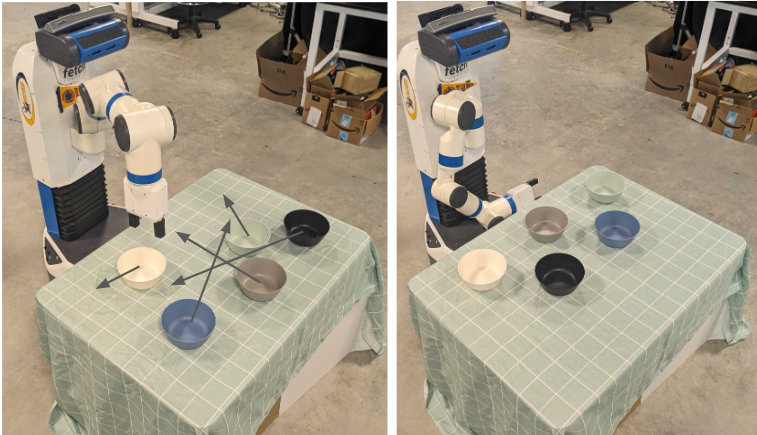}
    \caption{An example of mobile robot tabletop rearrangement, while black arrows are the desired rearrangement goals.}
    \label{fig:rearrangement_example}
\end{figure}

This paper presents an improved A*-based algorithm, Scalable Tabletop Rearrangement A*-based Planning (STRAP), designed to address the scalability concern. Unlike ORLA~\cite{orla}, which performs lazy evaluation during the expansion of the search tree, STRAP conducts lazy evaluation after each exploration step. That is, it utilizes the lazy evaluation in quick searching for feasible rearrangement plans from the current explored state to the goal state in each iteration. Consequently, while ORLA* may take longer to return a solution, STRAP quickly provides a high-quality solution. Additionally, STRAP incorporates the robot status to represent the system state, allowing exploration of more configurations and ultimately leading to improved planning outcomes compared to ORLA*~\cite{orla}. The evaluation section of this paper presents a series of experiments illustrating the enhancements provided by STRAP for both stationary and mobile
robots.

The paper is organized with related work in Section~\ref{related_works}, a problem statement in Section~\ref{problem_statement}, the improved algorithm in Section~\ref{methodology}. Then, we evaluate the algorithm with both stationary and mobile manipulators in Section~\ref{evalution}.

\section{Related works}\label{related_works}
\subsection{Tabletop Rearrangement}
The tabletop rearrangement task is a common manipulation problem involving a sequence of pick-and-place actions to arrange multiple objects into specific goal positions. The primary challenge arises from the dependencies between objects caused by potential collisions. To address this, a dependency graph is typically used, as described in~\cite{krontiris2015rss, krontiris2016icra}. Utilizing this graph transforms the problem into the feedback vertex set problem and the traveling salesperson problem, both of which are NP-hard, as noted by~\cite{han2018ijrr}, and causes the scalability problem. By incorporating an external buffer, ~\cite{han2018ijrr} were able to search for the minimal cost solution to complete the rearrangement task efficiently. However, it is common for the external buffer to be limited in size. To address this constraint, Running Buffer~\cite{runningbuffer} introduced the concept of a running buffer, which restricts the size of the external buffer while still enabling the completion of the rearrangement task efficiently.

\subsection{Rearrangement Planning with Internal Buffer}
Nevertheless, in many cases, external buffers are not available for single manipulators. Allocation of an internal buffer that ensures optimality is a computationally intensive task. To address this challenge, ~\cite{qureshi2021rss} proposed the Neural Rearrangement Planner, which employs deep learning techniques to solve the problem recursively. However, this approach still does not guarantee a high success rate, even when dealing with a small number of objects. On the other hand, pre-identifying potential buffer candidates~\cite{van2009rss, wang2020tase} can be a viable solution. However, this approach suffers from scalability issues in dense environments. Consequently, TRLB~\cite{trlb} aims to solve the rearrangement task using only internal buffers, even when dealing with many objects. Despite its advantages, this method does not take into account the travel cost. To further enhance this approach, Monte-Carlo Tree Search (MCTS) based methods~\cite{labbe2020ral, Baichuan2024icra} have been proposed; However, these methods are specifically designed for stationary robots and require an exceptionally long time to achieve high-quality solutions. Subsequently, ORLA*~\cite{orla} incorporates lazy buffer allocation within the A* search algorithm and prioritizes buffer poses based on various cost function optimizations for both stationary and mobile robots. Besides that, ORLA*~\cite{orla}  has demonstrated superior performance in terms of plan quality compared to MCTS-based methods.

To accomplish this, ORLA* classifies the states explored as deterministic (DS) and non-deterministic states (NDS). As the names suggest, deterministic states have well-defined object positions. These states have the objects either in the start states or the goal states. On the other hand, non-deterministic states occur as a result of an object being placed in a buffer position. Due to lazy buffer allocation, the buffer positions remain undefined until the exploration algorithm reaches a deterministic state. Upon reaching a deterministic state, the buffer positions are calculated by backtracking the actions. This way of calculation saves time by calculating buffer positions only when needed.  However, this design requires a state representation that considers only object arrangements, excluding robot locations. As a result, the state representation is incomplete, leading to suboptimal solutions. Furthermore, although ORLA* incorporates actions to stack objects on top of one another using a neural network to predict stability, our focus is solely on the rearrangement planning aspect. We exclude stacking actions to avoid unnecessary complexity.

\begin{figure}
    \centering
    \includegraphics[width=0.3\textwidth]{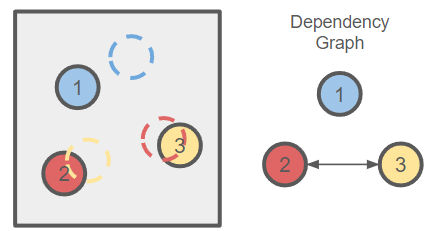}
    \caption{Left: A rearrangement task involves a table as the boundary box. The colored circles represent objects in their initial positions, while the dotted circles show their target positions. Right: Its dependency graph. }
    \label{fig:dependency_grasph}
\end{figure}

\section{Problem Statement}\label{problem_statement}

Given a 2D table workspace with $n$ objects, an arrangement 
$A$ is defined as $\{p_1, p_2, ..., p_n\}$, where each $p_i$ represents the pose of an object $i$. An arrangement is considered valid only if no objects collide with each other or with the boundaries of the workspace. Unlike previous study~\cite{orla}, we assume that objects are placed directly on the table without stacking. A rearrangement plan consists of a sequence of pick-and-place actions $[a_1, a_2, ...]$ that transition objects from one location to another. Each action $a$ specifies which object to move, where to pick it up, and where to place it. An action is valid only if it does not cause any collisions upon placing the object.


Similar to \cite{orla}, the quality of a rearrangement plan is evaluated based on the total travel cost required to move all objects from the initial arrangement, $A_{init}$, to the goal arrangement, $A_{goal}$. A plan with a lower total cost is considered to be of higher quality, indicating a more efficient rearrangement process.
The travel cost is defined differently for stationary and mobile robots. For a stationary robot, the travel cost is the Euclidean distance traveled by the end-effector (EE). For a mobile robot, the travel cost is the Euclidean distance traveled by the mobile base (MB) around the table. Additionally, akin to~\cite{orla}, the mobile robot always moves to the closest point around the table to the pick or place location before performing the manipulation.

\section{Methodology}\label{methodology}
This section describes our new A*-based algorithm, named STRAP, aiming to overcome the scalability issue of ORLA* in TORO problem for both stationary and mobile robots.

In the A* algorithm, incomplete state representation can result in incorrect or suboptimal solutions. Therefore, unlike ORLA*, STRAP incorporates the robot's pose into the state representation to ensure higher completeness. That is, given $n$ objects, a state $s$ is defined as follows:
\[s = \{p_r, p_1, p_2, ...\}\]
where $p_r$ is the robot's pose, while each $p_i$ is the pose of object i. On the other hand, each pick-and-place action $a$ is defined as
\[a = (k, p_{pick}, p_{place})\]
where $k$ is the object, $p_{pick}$ is the pick location, and $p_{place}$ is the place location.

Given a state, the action decision policy involves selecting which object that is not in its goal position to be pick-and-placed. Similar to ~\cite{orla, trlb}, the placement location of an action is determined based on the object's condition as following:
\begin{enumerate} \label{policy}
  \item If the object's goal position is available, then move it to its goal position.
  \item If the object's goal position is not available, then move it to a valid buffer location different from where it was.
\end{enumerate}

Starting from the initial state $s_{start}$, defined by both the initial arrangement and the robot's starting position, STRAP explores one state at a time, adding all subsequent states derived from the explored state to the priority queue, often called open list. This queue is sorted based on the $g$ and $h$ values, where $g(s)$ is the cost from the initial state to $s$, and $h(s)$ is the estimated cost from $s$ to the goal. In the next iteration, the priority queue pops the state with the lowest $f(s) = g(s) + h(s)$ and explores it until the explored state contains the goal arrangement or timeout.

To address scalability issues, STRAP is designed to deliver a solution within a short time frame. At the end of each iteration, STRAP performs a rapid goal-attempting procedure to identify a feasible, though not necessarily optimal, rearrangement plan from the current explored state to the goal arrangement. The best rearrangement plan, which minimizes the cost from the start state through the explored state to the goal state, is recorded. When a timeout occurs or the explored state matches the goal arrangement, the best plan available at that point is returned.

The following section covers state cost estimation, search tree exploration, goal attempting, and plan refinement.

\subsection{Cost Estimation}
To use the A* algorithm for the TORO problem, both $g(s)$ and $h(s)$ cost estimations must be precisely defined for each state $s$. Given a state $s$, $g(s)$ is the sum of the travel cost and manipulation cost used from $s_{start}$ to $s$. Specifically, given a pick-and-place action sequence $[a_1, a_2, ..., a_m]$ used to rearrange from $s_{start}$ to $s$, we have
\[g(s) =  MC \cdot m + TC(p_{robot}, p^{a_1}_{pick}) + TC(p^{a_1}_{pick}, p^{a_1}_{place})\]
\[+ \sum_{i=2}^{m}TC(p^{a_{i-1}}_{place}, p^{a_i}_{pick}) + TC(p^{a_i}_{pick}, p^{a_i}_{place})\]
Here, $p_{robot}$ denotes the initial robot location, $p^a_{pick}$ and $p^a_{place}$ are pick-and-place location of action $a$ respectively, $m$ is the number of actions, and $TC$ refers to the travel cost between two points. The $MC$ represents the manipulation cost defined by user for each pick-and-place action. For instance, $MC$ could account for the cost of opening and closing the gripper.

On the other hand, the $h(s)$ is the sum of manipulation cost times the number of objects not in goal and the total pick-and-place travel cost as the lower boundary cost. That is, given a state $s$ with a set of objects $[o_1, o_2, ..., o_l]$ which are not in their goal location, 
\[h(s) = MC \cdot l + \sum^{l}_{i=1} TC(p_i, p^g_i)\]
where $p_i$ is the current location of $o_i$, while $p^g_i$ is the goal location of $o_i$, and $l$ is the number of missing placed objects.

\begin{figure}
    \centering
    \includegraphics[width=0.25\textwidth]{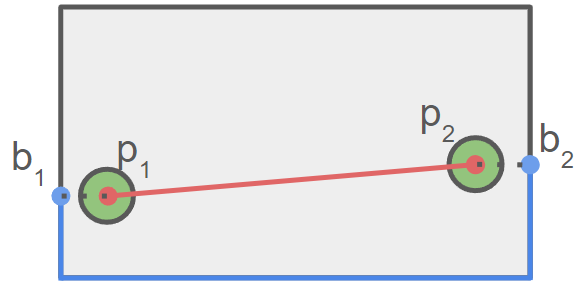}
    \caption{
    Different definitions of travel cost bewtween $p_1$ and $p_2$. For stationary robots, the $TC$ is the red line length. For a mobile robot, given $b_1$ and $b_2$ which are the closest points to $p_1$ and $p_2$ on the table side, the $TC$ is the traveling distance moving along the table side (blue). 
    }
    \label{fig:travel_cost}
\end{figure}

Additionally, the function of travel cost is calculated differently for stationary and mobile robots, as described in \cite{orla} and illustrated in Fig.~\ref{fig:travel_cost}. Given two object locations $p_1$ and $p_2$, for a stationary robot, $TC(p_1, p_2)$ is the Euclidean travel distance of the end-effector in the x-y plane. On the other hand, for a mobile robot, $TC(p_1, p_2)$ is the Euclidean travel distance to move along the table side from $b_1$ to $b_2$ where they are the closest points on the table side for the $p_1$ and $p_2$, respectively.

\subsection{Exploration}
Following the A* principle, each iteration explores the state $s$ with the lowest $f$ value from the priority queue. First, STRAP examines all states derived from $s$ based on the decision policy, calculating their corresponding $f$ values, then adds them to the priority queue. If a duplicate state is found in the queue, only the version with the lower $f$ value is retained. Additionally, the explored state $s$ is incorporated into the search tree as closed list in A*. 

When exploring new states from $s$, each object can lead to one or more possible states. If the object can be moved to its goal location, the new state will reflect the arrangement with the robot location after the robot picks and places the object at its goal. If its goal location is occupied, we sample multiple buffer locations on the table for it. If a buffer location is successfully allocated, the new state will reflect the arrangement with the robot location after the object is placed in the buffer. The number of sampling attempts is based on the number of objects in our implementation. For instance, in a state with three objects, where one is already at its goal location, another can be moved directly to its goal, and the third's goal location is occupied, the outcomes are as follows: The object at its goal location does not lead to any new states. The object that can be moved directly to its goal location leads to one new state. For the object that cannot be moved to its goal location, three random positions are sampled on the table. If two of these sampled positions are collision-free with other objects in the current state and suitable as buffer locations, this object results in two new states. Thus, this scenario produces a total of three subsequent states from the given state.

However, this approach can result in redundant actions, such as moving the same object twice in succession or moving an object from one buffer location to another when it could have been placed there directly in the first instance. To address this, two additional rules are implemented during exploration. First, during exploring from a given state to others, STRAP avoids exploring actions that involve moving the same object that was used to reach the given state. Second, to avoid moving an object from one buffer to another where it could have been placed directly initially, during sampling buffer location for an object, STRAP does randomly sampling on table if that object is never moved at all. If that object has been moved and is currently in a buffer location, STRAP only samples new buffer locations from areas where that object could not have been placed before being moved to its current buffer due to collision.
That is, in a given state $s_{current}$, when sampling a new buffer location for an object $o_{in\_buffer}$ already in a buffer, STRAP identifies the ancestor state $s_{ancestor}$ of $s_{current}$ in the search tree where $o_{in\_buffer}$ was last moved to its current buffer. The new buffer location is then sampled in a way that $o_{in\_buffer}$ would collide with other objects in the arrangement of $s_{ancestor}$. If this new sampled buffer location is collision-free with other objects in the arrangement of $s_{current}$, then it is considered as the new valid buffer location of $o_{in\_buffer}$.

\begin{figure}
\centering
\includegraphics[width=0.46\textwidth]{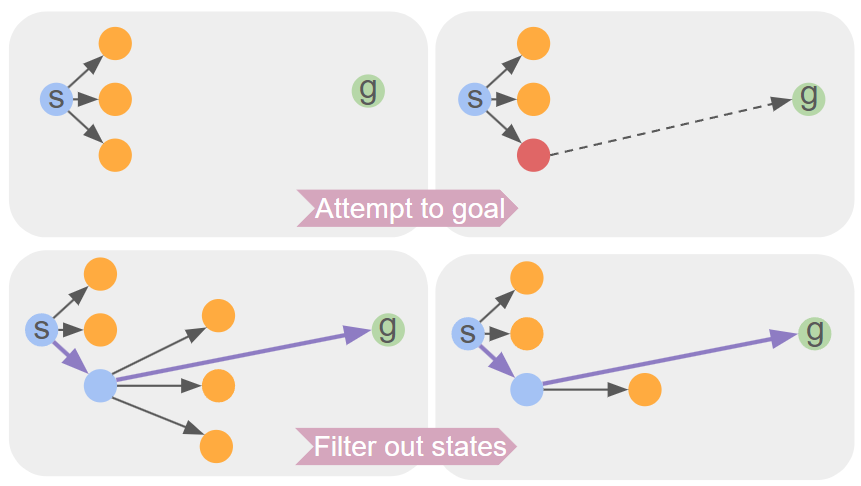}
\caption{Exploration process. The orange circles represent the states that are ready to be explored as in priority queue, whereas the blue circles indicate the states that have already been explored, and the green circle with $g$ is the state whose arrangement is the goal arrangement. In the upper section, during the exploration phase, the node with the minimum $f$ value is selected from the priority queue. The process of goal attempting involves searching for a feasible rearrangement solution from an explored state to the goal arrangement. In the lower section, if a plan exists that connects the start state to an explored state with a feasible solution to the goal, shown as purple path, and this plan is currently the best solution, its cost will be calculated. Subsequently, nodes in the priority queue with $f$ values exceeding this cost will be filtered out.
 }
 \label{fig:exploration_step}
\end{figure}

\subsection{Goal Attempting}\label{subsection:goal_attempting}
To ensure a solution within a short time frame, STRAP implements a process called \textit{goal attempting}, as shown in Fig.~\ref{fig:exploration_step}. After exploration in each iteration, it finds a feasible rearrangement plan from the current explored state to a state with goal arrangement. Since the process is executed at every iteration, it must be fast. To achieve this efficiency, STRAP leverages a state-of-the-art rearrangement local solver with lazy buffer allocation from ~\cite{trlb}, which is optimized for solving rearrangement tasks in cluttered environments in short time, even if the resulting plan is not optimal. After each goal attempting, the current best plan is re-evaluated by concatenating the action sequence leading to the explored state with the newly generated plan to the goal state. If this combined plan has a lower rearranging cost than the previous best plan, it is adopted as the new best plan. Consequently, states in the priority queue with an $f$ value exceeding the cost of the new best plan are filtered out, thereby streamlining the process and reducing unnecessary sorting operations in the priority queue.

\begin{algorithm}
\caption{PlanRefinement}\label{alg:plan_refinement}
\begin{flushleft}
\textbf{INPUT:} \\ $actionSequence$ - pick-and-place action sequence\\
$arr_{start}$ - start arrangement: $\{1: p_1, 2: p_2, ...\}$
\end{flushleft}
\begin{algorithmic}[1]
\State $B\leftarrow\{\}$
\State $H \leftarrow \{0: arr_{start}\}$ \Comment{Arrangement history}
\State $arr_{current} = arr_{start}$
\For{$i, action \in \text{Enum}(actionSequence)$} \Comment{start at 1}
\State $k, p_{pick}, p_{place} = action$
\If {$k\in B$} \Comment{if object k was moved}
\State $ToBuf = B[k]$ \Comment{previous action index on k}
\State $C = getConstraint(H[ToBuf] / k)$
    \For{$a \in actionSequence[ToBuf: i]$}
        \State $C.addConstraint(a[2])$
    \EndFor
\State $p_1 = actionSequence[ToBuf][1]$
\State $p_2 = actionSequence[i-1][2]$
\State $p_3 = actionSequence[ToBuf+1][1]$
\State $p_4 = actionSequence[i][2]$
\State $P = BufGen(C)$ 
\State \Comment{Generate a buffer set under constraint C }
\State $p^* \leftarrow$ select from $P$ min $\sum\limits^4_{j=1} TravelCost(p^*,p_j)$
\State $actionSequence[ToBuf][2] = p^*$
\State $actionSequence[i][1] = p^*$
\EndIf
\State $B[k] = i$
\State $arr_{current}[k] = p_{place}$ \Comment{update arrangement}
\State $H[1] = arr_{current}$ \Comment{record arrangement}
\EndFor
\end{algorithmic}
\end{algorithm}
\subsection{Rearrangement Plan Refinement}
After termination, the current best rearrangement plan should be refined further. Given a pick-and-place action sequence
$[a_1, a_2, ..., a_m]$,
our refining process iteratively improves the plan until the cost no longer improves.

In the refinement process, the algorithm iterates over all the actions involving moving an object to a buffer location, called ``to-buffer'', from the beginning of the sequence. Each iteration aims to find a better buffer location for ``to-buffer'' actions. Once an improved buffer location is identified, the object's location is updated, and the subsequent ``to-buffer'' actions are optimized with the modified placement.

For a to-buffer action, the following four locations are important:
\begin{description}
  \item[$p_1$]Last gripper position before placing at buffer.
  \item[$p_2$]Last gripper position before picking at the buffer.
  \item[$p_3$]First gripper position after placing at the buffer.
  \item[$p_4$]First gripper position after picking at the buffer.
\end{description}

From the moment the object is moved into the buffer until it is moved out, we can search for a collision-free buffer location during this interval and minimize the total travel cost from this buffer location to all four positions using sampling. This iteration process is demonstrated as the pseudocode in Alg.~\ref{alg:plan_refinement}.

This paragraph provides more details of the refinement. Given an action sequence and initial arrangement, the goal is to optimize the buffer location. First, we initialize a dictionary $B$ that contains the action index in the action sequence for each object moved previously. Additionally, $H$ represents the arrangement history. Starting from line 4, each action in the sequence is processed. If the object has been moved before, it indicates it was placed in a buffer location since objects are not moved from their goal positions. From lines 6 to 7, if an object $k$ was moved previously, we identify the action index $ToBuf$ where it was moved to a buffer location. From lines 8 to 11, we determine the obstacle constraints to ensure the buffer location does not collide from action $ToBuf$ to the current index. From lines 12 to 18, based on these constraints, we sample a set of possible buffer locations and evaluate the best one. Finally, from lines 19 to 20, the action sequence is updated accordingly.

\section{Evaluation}\label{evalution}
To ensure a fair comparison, we applied the disk test cases from TRLB~\cite{trlb} on both stationary and mobile settings across the following algorithms:
\begin{enumerate}
  \item STRAP: Planner presented in this paper.
  \item ORLA*: Planner based on~\cite{orla} as the
        state-of-art.
  \item TRLB: Planner based on~\cite{trlb} which doesn't return high-quality solutions.
  \item Iterative TRLB: Runs TRLB algorithm on the same task multiple times until a timeout, then returns the best one.
  \item MCTS: Uses Monte Carlo tree search with the reward function from~\cite{Baichuan2024icra} and including robot configuration in the state.
\end{enumerate}

\begin{figure}
    \centering    
    \includegraphics[width=0.5\textwidth]{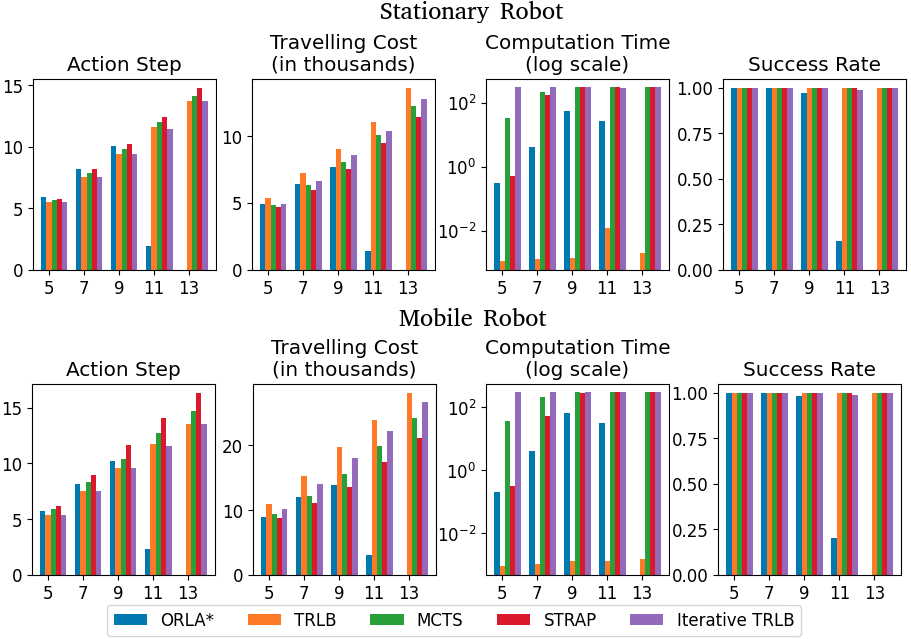}
    \caption{The comparisons of different algorithms for stationary \& mobile robots on different object number. The horizontal axis represents object number.}
    \label{fig:result1}
\end{figure}

\subsection{Results}

Our evaluation uses the same metrics as in~\cite{orla}: average number of actions, average travel cost, average computation time, and success rate, while manipulation cost is one.

As illustrated in Fig.~\ref{fig:result1}, we compare the general performance of all algorithms with 5 minutes timeout and 0.2 density, and it reveals the following trends: ORLA* struggles to scale effectively with increasing object number, regardless of whether the robots are stationary or mobile, whereas the other algorithms consistently maintain a high success rate. Regarding travel cost, both ORLA* and STRAP outperform others. ORLA* only occasionally shows very low cost in scenarios where its success rate is low.

To better understand the time consumption of ORLA*, we conducted tests with a one-hour timeout on 70 tasks in both mobile and stationary settings, each involving 11 objects. Our results show that in the mobile setting, 25\% of the cases required more than 2500 seconds to execute, while in the stationary setting, 75\% of the cases exceeded 1400 seconds.

\begin{figure}
    \centering    
    \includegraphics[width=0.47\textwidth]{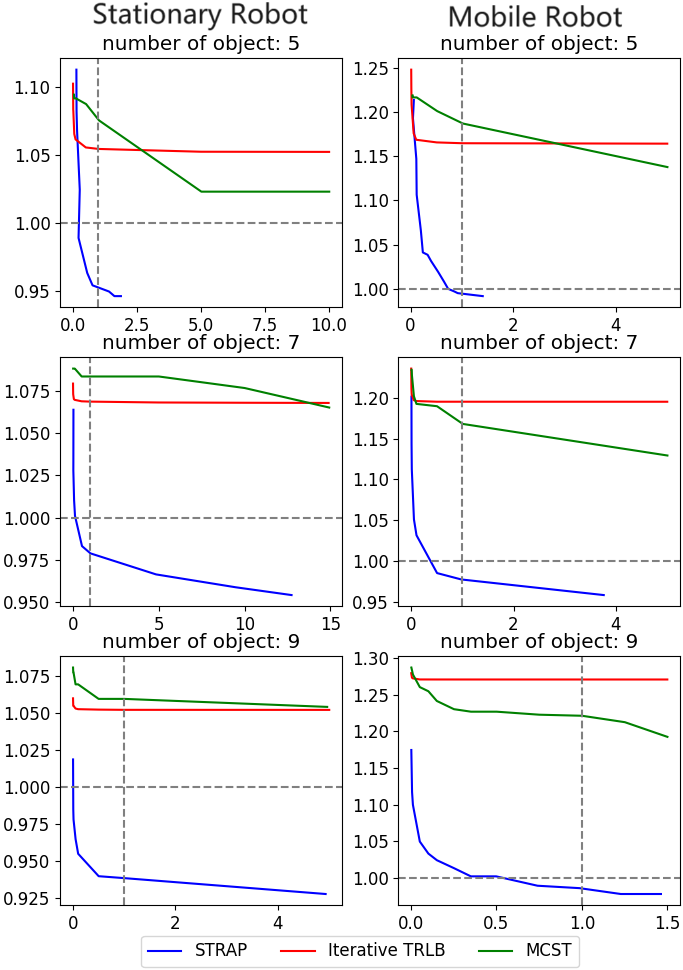}
    \caption{The travel cost ratio compared to ORLA* over different timeout ratio of ORLA* on different number of objects for both the mobile and the stationary settings. The STRAP may stop early if the priority queue runs out of states.}
    \label{fig:over_time_result}
\end{figure}

\begin{figure}
    \centering    \includegraphics[width=0.5\textwidth]{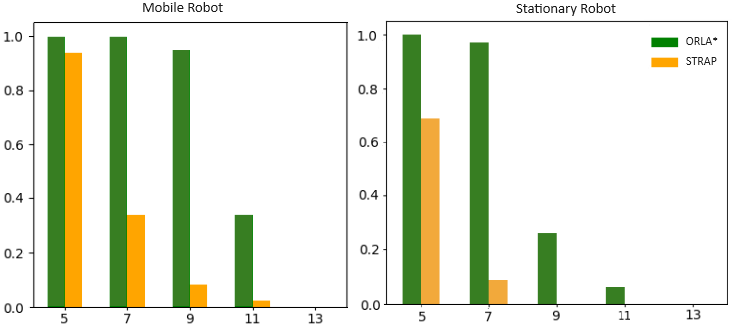}
    \caption{Success rate of ORLA* and STRAP without goal attempting as a function of object number.}
    \label{fig:no_goal_attempt}
\end{figure}

Furthermore, we compare the travel costs of various algorithms over time against ORLA* for both mobile and stationary settings, as shown in Fig.~\ref{fig:over_time_result}. This evaluation is restricted to scenarios (5, 7, 9 objects) where all algorithms achieve a high enough success rate. Given 30 tasks, we first determine the average computation time of ORLA* across different numbers of objects in both mobile and stationary settings. We then evaluate other algorithms on these same 30 tasks using different timeout ratios of ORLA*. As shown in Fig.~\ref{fig:over_time_result}, neither MCTS nor Iterative TRLB achieves a lower travel cost compared to ORLA* in reasonable time. Conversely, STRAP matches the travel cost of ORLA* while requiring significantly less computation time. Additionally, STRAP can lower travel costs more than ORLA* while using the same computation time. This is because STRAP includes the robot location in its state representation, making its state representation more complete than ORLA*, which does not consider the robot location.

\subsection{Ablation Study}

\begin{figure}
    \centering    \includegraphics[width=0.5\textwidth]{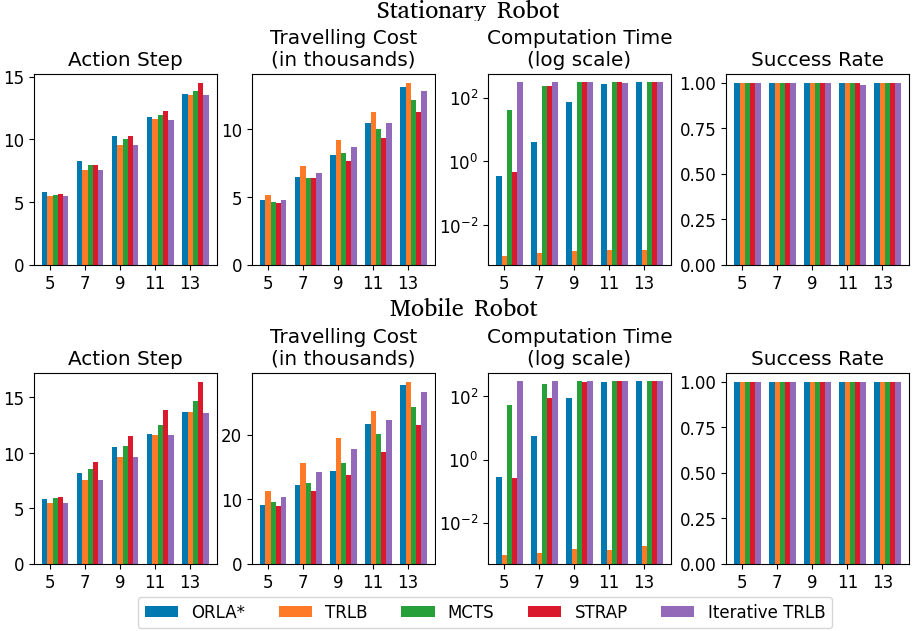}
    \caption{The comparison where both ORLA* \& STRAP with goal attempting.}
    \label{fig:both_with_goal_attempt}
\end{figure}

This section performs two ablation studies. First, we compare ORLA* and STRAP without goal attempting to verify whether lazy evaluation of ORLA* actually improves the performance as an A* based algorithm. Using 100 tasks with a timeout of 5 minutes,~Fig.~\ref{fig:no_goal_attempt} demonstrates that ORLA* achieves a higher success rate than STRAP without goal attempting, indicating that lazy evaluation indeed benefits A*-based algorithms.

Next, we examine whether incorporating goal attempting into ORLA* can further improve its performance. In ORLA*, goal attempting can only be executed in the deterministic states due to the nature of the algorithm. As shown in Fig~\ref{fig:both_with_goal_attempt}, the success rate of ORLA* is constantly high. However, ORLA*'s travel cost does not improve and be better than STRAP. The reason for this behavior is that there are very few deterministic states in ORLA*'s search tree. 

\begin{figure}
    \centering    
    \includegraphics[width=0.45\textwidth]{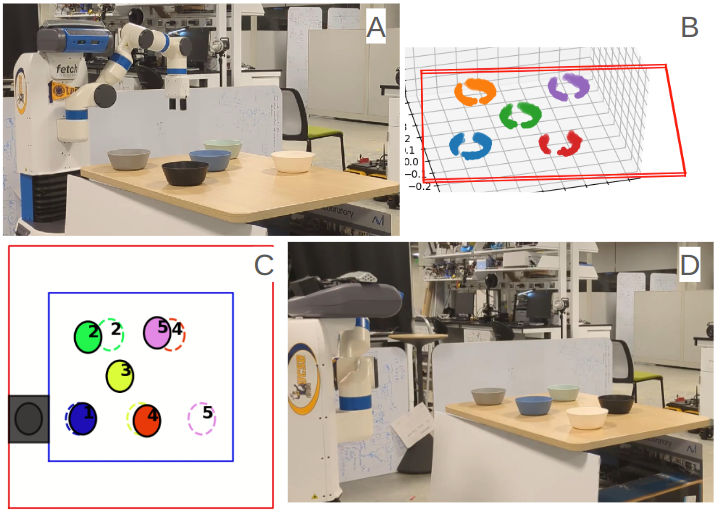}
    \caption{Real world experiment. (A) Given a set of bowls on the table. (B) Estimate each bowl's position in the table frame. (C) Generate the rearrangement plan. (D) Execute the plan on Fetch Robot.}
    \label{fig:real_world_exp}
\end{figure}

\subsection{Real World Experiment}
The accompanying video, as shown in Fig.~\ref{fig:real_world_exp}, demonstrates STRAP running on the Fetch robot~\cite{fetch2016} to rearrange bowls on the table. After the real-world experiment, we have multiple concerns. First, for the mobile robot, a lot of pick-and-place locations are close to each other, so certain mobile movements are not necessary. Therefore, in the future, this should be taken into account. Second, real-world rearrangement tasks are prone to failure if grasp estimation is inaccurate, highlighting the need for a recovery system to handle such failures.

\section{Conclusion}
In this work, we introduce STRAP for high-quality tabletop object rearrangement planning, designed for both stationary and mobile robots. Our proposed method addresses critical limitations of previous techniques, particularly in terms of scalability. Extensive evaluations have demonstrated that our algorithm not only scales more effectively with increasing problem complexity but also achieves higher levels of quality compared to ORLA*. Specifically, our approach adapts efficiently to more extensive and intricate scenarios, ensuring that the rearrangement solutions are both feasible and of high-quality.

\addtolength{\textheight}{-12cm}   




\section*{ACKNOWLEDGMENT}

We would like to express our deepest gratitude to Kai Gao, the author of ORLA*~\cite{orla} and TRLB~\cite{trlb}, for his invaluable assistance in helping us develop the ORLA* for this research. 


\end{document}